\documentclass[10pt,letterpaper]{article}
\usepackage[top=0.85in,left=2.75in,footskip=0.75in]{geometry}

\usepackage{amsmath,amssymb}

\usepackage{changepage}

\usepackage[utf8x]{inputenc}

\usepackage{textcomp,marvosym}

\usepackage{cite}

\usepackage{nameref,hyperref}

\usepackage{microtype}
\DisableLigatures[f]{encoding = *, family = * }

\usepackage[table]{xcolor}

\usepackage{array}

\newcolumntype{+}{!{\vrule width 2pt}}

\newlength\savedwidth

\usepackage{setspace} 
\raggedright
\usepackage[aboveskip=1pt,labelfont=bf,labelsep=period,justification=raggedright,singlelinecheck=off]{caption}

\makeatletter
\renewcommand{\@biblabel}[1]{\quad#1.}
\makeatother

\date{}

\usepackage{lastpage,fancyhdr,graphicx}
\usepackage{epstopdf}
\usepackage{csvsimple,longtable,booktabs}
\usepackage{subfigure}
\usepackage{float}
\usepackage{chngcntr}

\fancyheadoffset[L]{2.25in}
\fancyfootoffset[L]{2.25in}

\begin{document}
\vspace*{0.2in}

\begin{flushleft}
{\Large
\textbf\newline{Look Who's Talking: Bipartite Networks as Representations of a Topic Model of New Zealand Parliamentary Speeches} 
}
\newline
\\
B. Curran\textsuperscript{1}, 
K. Higham\textsuperscript{2}, 
E. Ortiz\textsuperscript{3}, 
D. Vasques Filho\textsuperscript{1}
\\
\bigskip
\textbf{1} Te P\={u}naha Matatini, University of Auckland, Private Bag 92019, Auckland 1011, New Zealand
\\
\textbf{2} Te P\={u}naha Matatini, Victoria University of Wellington, PO Box 600, Wellington 6140, New Zealand
\\
\textbf{3} Departament de F\'{\i}sica de la Mat\`eria Condensada, Universitat de Barcelona, Mart\'{\i} i Franqu\`es 1, 08028 Barcelona, Spain.

\smallskip

%
%
\Yinyang The authors contributed equally to this work.

* b.curran@auckland.ac.nz

\end{flushleft}
\section*{Abstract}

Quantitative methods to measure the participation to parliamentary debate and discourse of elected Members of Parliament (MPs) and the parties they belong to are lacking. This is an exploratory study in which we propose the development of a new approach for a quantitative analysis of such participation. 
We utilize the New Zealand government's digital Hansard database to construct a topic model of parliamentary speeches consisting of nearly 40 million words in the period 2003-2016. A Latent Dirichlet Allocation topic model is implemented in order to reveal the thematic structure of our set of documents. This generative statistical model enables the detection of major themes or \textit{topics} that are publicly discussed in the New Zealand parliament, as well as permitting their classification by MP. Information on topic proportions is subsequently analyzed using a combination of statistical methods. We observe patterns arising from time-series analysis of topic frequencies which can be related to specific social, economic and legislative events. We then construct a bipartite network representation, linking MPs to topics, for each of four parliamentary terms in this time frame. We build projected networks (onto the set of nodes represented by MPs) and proceed to the study of the dynamical changes of their topology, including community structure. 
By performing this longitudinal network analysis, we can observe the evolution of the New Zealand parliamentary topic network and its main parties in the period studied. 


\section{Introduction}
\label{sec:introduction}

Topic models have received widespread attention in recent years \cite{titov2008,ramage2010,zhao2011,grimmer2010} as they have proven to be useful tools for dealing with the vast amount of semantic information that is becoming available. Topic modeling is a set of machine learning techniques that take a collection of documents as input and attempts to discern the themes that pervade them \cite{blei2012}. However, the methods that topic models utilize to search, summarize and understand large electronic archives have rarely been applied to political texts. 

The New Zealand government has been making parliamentary transcripts ('Hansard') available in digital format since 2003. Suitable annotation of these transcripts allow them to be used as a corpus for the development of topic models. This comprehensive corpus of political text can then be examined through a number of lenses. Topic models allow us to monitor the ebb and flow of themes that are discussed in parliament over multiple years and associate particular themes with individual Members of Parliament (MPs). This allows the identification of trends of topics that particular parties follow.  That is, we may observe which issues are discussed repeatedly with great interest and which cease to be mentioned.

In the four parliamentary terms analyzed there was a transition of power from the $5^{\mathrm{th}}$ Labour government (1999-2008) to the $5^{\mathrm{th}}$ National government (2008-). The left-leaning Labour Party and right-leaning National Party have been the two parties sharing power for most of the $20^{\mathrm{th}}$ century. In 1996, the method of electing MPs was changed to a mixed-member proportional (MMP) system and the two major parties were joined by a number of smaller parties. These smaller parties have sometimes held the balance of power, with the left-wing Green Party as the largest of these.  

A number of textual analyses of political speeches are concerned with finding where on the political spectrum a speaker falls (e.g. \cite{laver2003,slapin2008,laver2000}). Topic modeling as applied in our analyses cannot determine the sentiment of a statement or speech. Despite this fact, multiple aspects of politicians' policy interests can be unraveled with further statistical analysis. 

Here we construct bipartite networks \cite{koskinen2012modelling,wasserman1994social,breiger1974duality,guillaume2006}, 
whereby sets of MPs are linked to a set of topics, with each link representing a topic that is of clear interest to a particular MP, based on the content of their parliamentary speeches. We can then decompose such bipartite networks into their two projections: the MP-projection and the topic-projection. The former represents a network where the links between MPs indicate the existence of a mutual interest, and the latter represents a network where links represent topics that co-occur as interests of a particular MP. In this study, we make use only of MP-projections. Measuring properties such as the node degree (i.e. the number of links that connect it to other nodes), homophily \cite{mcpherson2001birds,newman2003mixing} and clustering and community structure \cite{girvan2002community,newman2003structure} of these networks provides information about their underlying topology. For instance, one can discover whether or not the typical range of interests of an MP is changing, as well as patterns in this behavior over time. Moreover, we apply community-detection methods \cite{fruchterman1991graph,blodel2008fast} in order to find clusters of politicians that share interests, and investigate the partisan composition of these communities.

This work is of an exploratory nature, in that our goal is twofold: to present a novel quantitative approach of measuring political activity and to demonstrate the benefits of performing quantitative analysis in a domain normally reserved for qualitative approaches, by using a combination of machine learning and complex networks techniques.

The remainder of this paper is organized as follows: the Methods and Data sections \ref{sec:topic} and \ref{sec:bipartite} introduce fundamental aspects of topic modeling and bipartite networks respectively and outline the preparation and organization of our data; Sections \ref{sec:results} and \ref{sec:discussion} present the results of our analyses alongside a discussion and our conclusions. 

\section{Methods and Data}
\label{sec:methods}

\subsection{Data}
\label{sec:data}
The semantic data we are using for our analyses are extracted from the New Zealand Hansard database \cite{NZPD}. Hansard presents records of what is said in the debating chamber as debates (a collection of speeches on a particular topic), speeches (individual statements by MPs) or dailies and volumes (collections of speeches over different time periods). By considering only those documents labeled in that database as a `speech' we were able to find out in which topics specific MPs were engaging with. This makes it possible to associate speeches and by extension MPs with topics of interest over time. 

\begin{figure}[!h]
\centering
\includegraphics[width=0.8\textwidth]{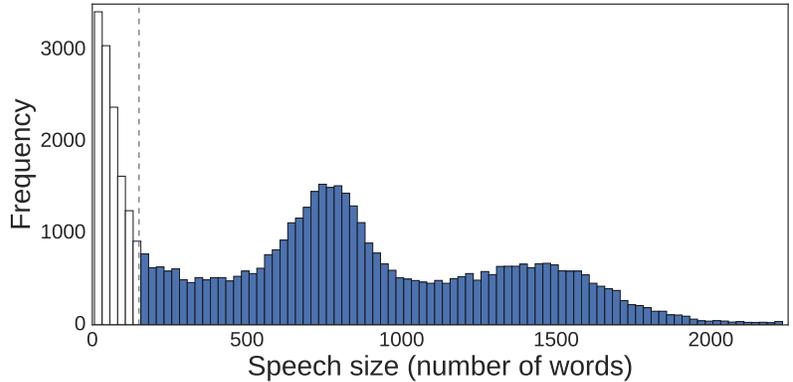}
\caption{Distribution of word-counts for New Zealand Parliamentary speech delivered in the period February 2003 to August 2016. In blue, the speech sizes used in our analysis. Speeches shorter than 150 words are omitted from our analysis due to lack of topical content, this threshold is indicated by the dashed vertical line.\label{speechdist}}
\end{figure}

Once these data are obtained, we observe that many speeches are rather short and contain little topical content. An example is given below, which comes from a committee discussion on the \textit{Shop Trading Hours Amendment Bill} and was published in Hansard Volume 716 on the $17^{\mathrm{th}}$ of August 2016 \cite{NZPD}:
\newline
\newline
"\textbf{CHAIRPERSON (Lindsay Tisch):} Just a point: this debate concludes at quarter past and to whoever is speaking at the time, I will be stopping it at that point."
\newline
\newline
In an attempt to remove these non-topical speeches, we have removed from our database speeches with 150 words or less, which constitute about 20\% of the database. This cut-off is shown in Fig. \ref{speechdist} which presents the distribution of word-counts per speech. This decision is informed by observations of the insufficient topical content of speeches below this threshold. \\

\subsection{Topic Models} 
\label{sec:topic}
The process of topic modeling involves utilizing a set of algorithms that have been developed to understand the underlying thematic structure of a corpus. The simplest and most commonly used topic model is Latent Dirichlet Allocation (LDA) \cite{campbell2014latent}. Within the framework of LDA, each document is a mixture of corpus-wide topics and each topic can be understood as a distribution over keywords. The total number of documents comprising the corpus is denoted as $D$ and the total number of topics as $K$. Additionally, the order of words that comprise the document is not considered, only the frequency with which words appear. 

From the perspective of LDA, documents are imagined to be the result of a generative process. This is the process by which the model assumes the documents arose given certain \textit{hidden variables}. The word-distributions per document are observed, while the topic structure -- per-word topic assignment, per-document topic proportions and per-corpus topic distributions -- are hidden elements. Therefore, the central computational problem for LDA is to infer the hidden structure that likely generated the observed corpus.
This means computing the conditional distribution of the hidden variables given what is observed. This conditional distribution is usually referred to as the \textit{posterior} and can be expressed as 
\begin{equation}
p(\beta_{1:K},\theta_{1:D},z_{1:D}|w_{1:D})=\frac{p(\beta_{1:K},\theta_{1:D},z_{1:D},w_{1:D})}{p(w_{1:D})}
\end{equation}
where $\beta_{1:K}$ are all topics in the corpus, $\theta_{1:D}$ are the per-corpus topic proportions, $z_{1:D}$  the per-corpus topic assignments and $w_{1:D}$ the whole set of observed words. Unfortunately, computing the posterior is computationally unfeasible and hence needs to be approximated by an inference algorithm. Consequently, topic modeling algorithms are commonly classified as sampling-based algorithms or variational algorithms \cite{blei2003}. 

In this work, we used the MALLET software package \cite{mccallum2002} for the topic modeling component of our analysis. MALLET implements Gibbs sampling \cite{griffiths2002}, which constructs a sequence of random variables in a Markov chain, where each variable is dependent on the previous. The algorithm then assumes that the true posterior distribution is the limiting distribution of this sequence, and obtains an approximation to this posterior using these samples. For a full mathematical description of LDA and a further discussion of the methods used to estimate a posterior, see \cite{blei2003}.\\

\begin{figure}[!h]
\vspace{0cm}
\centering
\includegraphics[width=0.7\textwidth]{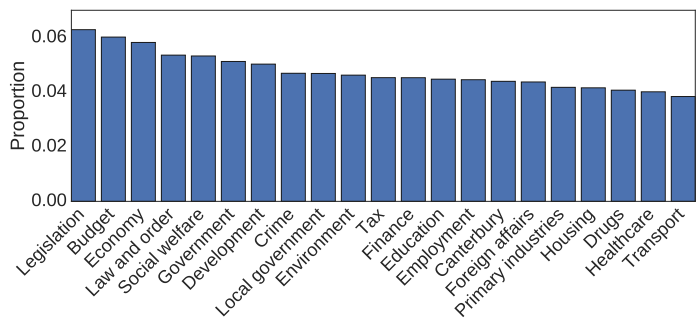}
\vspace{0mm}
\caption{Topics identified by an LDA-based topic model of the Hansard speech corpus for the period February 2003 to August 2016 and their respective normalized proportions as a fraction of all the topical content discovered after the omission of nine topics with little specific thematic content. A list of topic and the key words that comprise them can be found in Table \ref{keywordstable} in Supporting Information.} 
\label{fig:topicsIdentified}
\end{figure}

LDA assumes the topics are the same for all documents, and only the topic proportions vary. Therefore, MALLET requires an input which specifies the number of topics to be discovered. Choosing this number is critical to the success of a topic model, as too few topics may merge distinct themes, while too many topics may introduce many "themes" consisting of vocabularies that appear to have nothing in common, or even start splitting topics that were identifiable at smaller input values. For our analysis it is important that the topics are easily identifiable and distinct from one another. We found that 30 topics satisfies these requirements. Identified topics and their corresponding keywords can be found in Table \ref{keywordstable}. It is worth noting, however, that some topics (nine of them, corresponding to about 36\% of the corpus) appeared to consist mainly of terms that were primarily either procedural or general rhetoric, such as "proud", "hope" or "nation". As this language reveals little in the way of substantive interactions, such topics were omitted from our subsequent analyses after networks had been inferred. Fig. \ref{fig:topicsIdentified} shows the remaining topics with their rescaled proportions.\\

\subsection{Bipartite networks}
\label{sec:bipartite}

A bipartite network is mathematically defined as a graph $G = \{U,V,E\}$, where $U$ and $V$ are disjoint sets of nodes and $E = \{(u_{i},v_{j}):u_{i} \in U, v_{j} \in V\}$ is the set of links connecting these nodes. For our purposes, the sets $U$ and $V$ correspond to the sets of MPs and topics, and the set $E$ represents the links that emerge when an MP speaks sufficiently frequently about a topic. No connections among nodes of the same set are allowed in the bipartite network, that is, MPs are connected only to topics not others MPs and vice versa. Each set of nodes can have independent properties, such as the probability distribution for their nodes degree, or the number of nodes (system size).

Once we find the set of topics that a particular MP speaks about `often' enough (this criterion is defined below), these are represented as links between the MP and those topics. After this process is completed for all MPs, we can construct a bipartite network where nodes representing MPs are connected only to nodes representing topics, and vice versa. 

Bipartite structures play an important role in the analysis of social and economic networks. They are normally used to represent conceptual relations - such as membership, affiliation, collaboration, employment, ownership and others - between two different types of entities within a system \cite{koskinen2012modelling,wasserman1994social,breiger1974duality}. Often, we are more interested in one of the types of nodes (e.g. MPs) and, in order to investigate the relationships between them, we create a new network with only these nodes. This new graph is a \textit{projection} of the original bipartite network.

\begin{figure}[!h]
\centering
\includegraphics[width=0.4\textwidth]{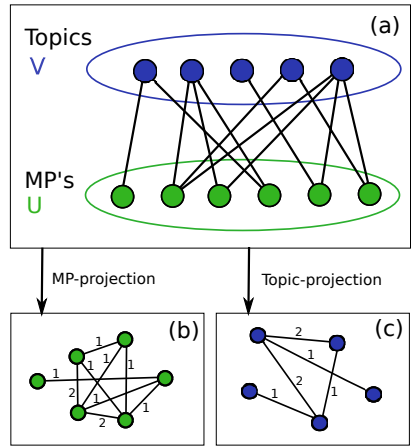}
\vspace{0.3cm}
\caption{Schematic drawing of the (a) bipartite network and its projections onto set of nodes (b) $U$, where nodes represent MPs, and (c) $V$, where nodes represent topics. In (b) and (c) the numbers on top of the links indicate the weights, which correspond to the number of common neighbors that two linked nodes share in the bipartite network. }
\label{bipnetwork}
\end{figure}

Topic modeling results in a natural bipartite network with projections that can be easily interpreted. The projections of a bipartite network are obtained by connecting nodes which share a common neighbor. That is, if two MPs are both linked to the same topic in the bipartite network, then they are linked in the MP-projection. For a bipartite network, this process results in two completely separated components, each composed exclusively of one type of node (MPs and topics in our case). Fig. \ref{bipnetwork} shows a schematic drawing of a bipartite network and its possible projections. The edges between nodes in these projections are then weighted, dependent on the number of neighbors the nodes share in the bipartite network. In our analyses we use \textit{simple weighting} method \cite{zhou2007bipartite}, whereby each edge has a weight that equals the number of neighbors the nodes share in the bipartite graph. If two MPs are linked to the same three topics in the bipartite graph, then the edge linking them in the MP-projection will have weight equal to three.

The weighting in these projections offers a way to eliminate edges that represent tenuous links. This is important, as \textit{complete subgraphs} (where every node is connected to every other node in the subgraph) of MPs are generated by every topic. This means that every MP that speaks about a popular topic is connected to every other MP that speaks about that same topic. The existence of popular topics can make analyses such as community detection challenging in the absence of weighting.

In order to build bipartite networks connecting MPs to topics, we looked at the corpus of each MPs speeches in more detail. We considered an MP to be connected to a topic when at least 6.7\% of the MP's speeches over the course of a year was assigned to that topic by MALLET. This occurs when MPs talk about a topic twice as much as would be expected if they were talking about all topics equally within a year. This method, removes topics that MPs only touch on briefly or in passing, which does not indicate engagement with the topic. Finally, MPs that had spoken less than $10^4$ words in the entire term were removed form the network for the lack of significance in the volume of words spoken.\\


\section{Results}
\label{sec:results}
\subsection{Words spoken}
\label{sec:wordsSpoken}
\begin{figure}[ht]
		\centering
		\includegraphics[width=0.75\textwidth]{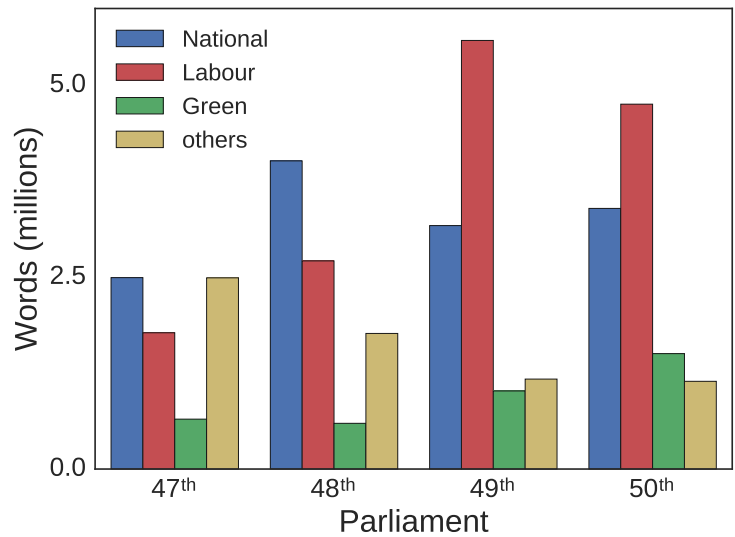}
		\caption{Total number of words spoken by each party in their speeches longer than 150 words, for each of the four complete parliamentary terms studied.}
		\label{wordsparty}
\end{figure}

Despite having fewer MPs, opposition parties tend to have a greater total word count than the governing party. Figure \ref{wordsparty} shows the total word count for each of the 3 largest parties (as of the 50th parliament) over the course of 4 parliaments. In each parliament, the total word count for opposition parties exceeds that of the governing party. The increase in words spoken does not appear to be driven by any particular MP or small group of MPs (see Supporting Information, Fig. (\ref{fig:wordsbyMP}).

\subsection{Time Series of Topic Popularity}
\label{sec:topicProportions}

\begin{figure}[!h]
\vspace{0cm}
\centering
\includegraphics[width=0.7\textwidth]{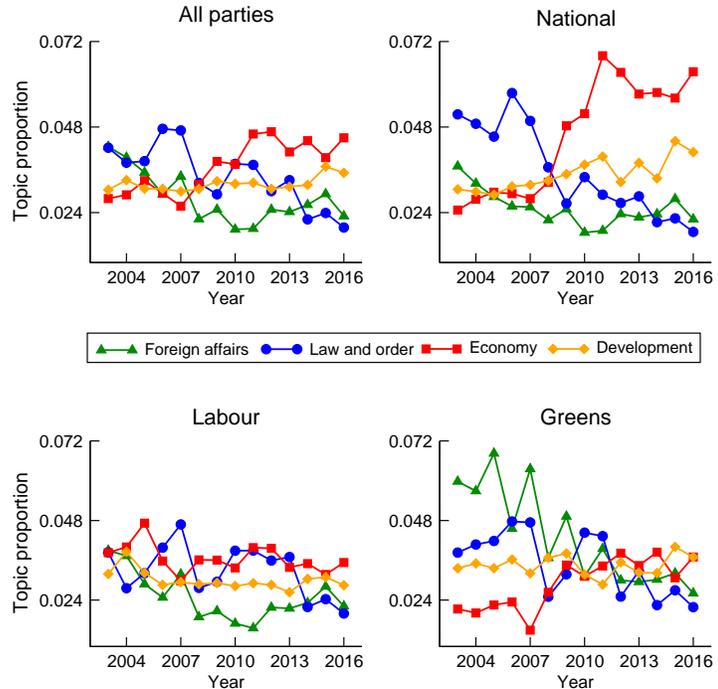}
\vspace{0mm}
\caption{Evolution over time of the proportions in which the topics labeled as  \textit{Foreign Affairs}, \textit{Law and order}, \textit{Economy}, and \textit{Development}, are discussed at the Parliament, and the corresponding decomposition by party.} 
\label{fig:timeevolutionbyparty1}
\end{figure}

\begin{figure}[!h]
\vspace{0cm}
\centering
\includegraphics[width=0.7\textwidth]{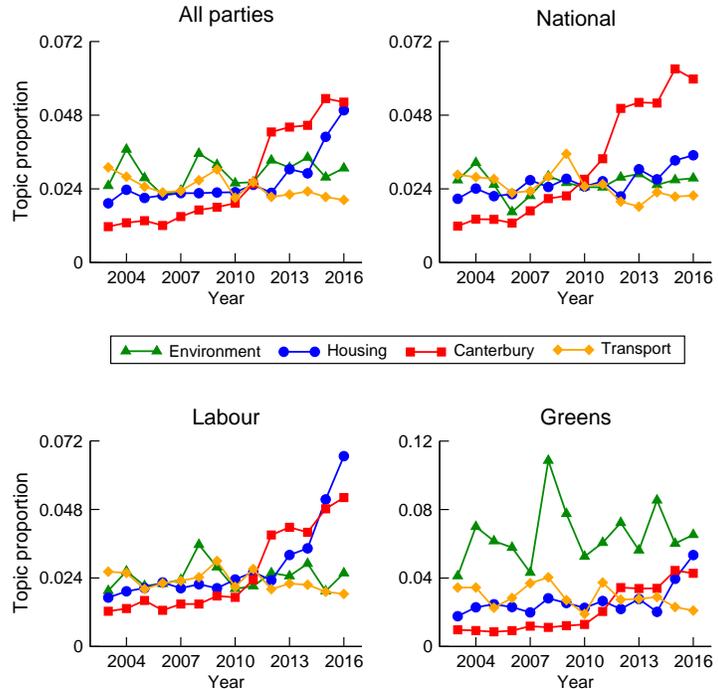}
\vspace{0mm}
\caption{Evolution over time of the proportions in which the topics labeled as  \textit{Environment}, \textit{Housing}, \textit{Canterbury}, and \textit{Transport}, are discussed at the Parliament, and the corresponding decomposition by party. Note the change in scale in the Topic proportion axis for the Green Party time series to accommodate their notable interest in environmental policy.} 
\label{fig:timeevolutionbyparty2}
\end{figure}

Allowing a decomposition by party, we ran a topic model on data concatenated by MP and year. The topic proportions obtained over a total number of 30 topics are normalized for each year so that they are comparable across a time span of 14 years. Proceeding this way, we can reproduce the evolution of topic popularity over time at the Parliament and its decomposition for each of the three most represented parties. Clear trends and differences across parties are visible in Fig. \ref{fig:timeevolutionbyparty1} and \ref{fig:timeevolutionbyparty2}. Evolution of proportion of other argued topics appears in Fig. \ref{fig:timeevolutionbyparty3}, Supporting Information.

Changes in topics being discussed often correlate with real world events such as the financial crisis of 2008 \ref{fig:timeevolutionbyparty2}), Canterbury and the Christchurch earthquakes of 2010/2011 (\ref{fig:timeevolutionbyparty1}) and more recently the housing crisis from about 2013 (\ref{fig:timeevolutionbyparty1}).   



\subsection{The Parliamentary Speech Network}

\begin{figure}[!h]
\centering
\includegraphics[width=\textwidth]{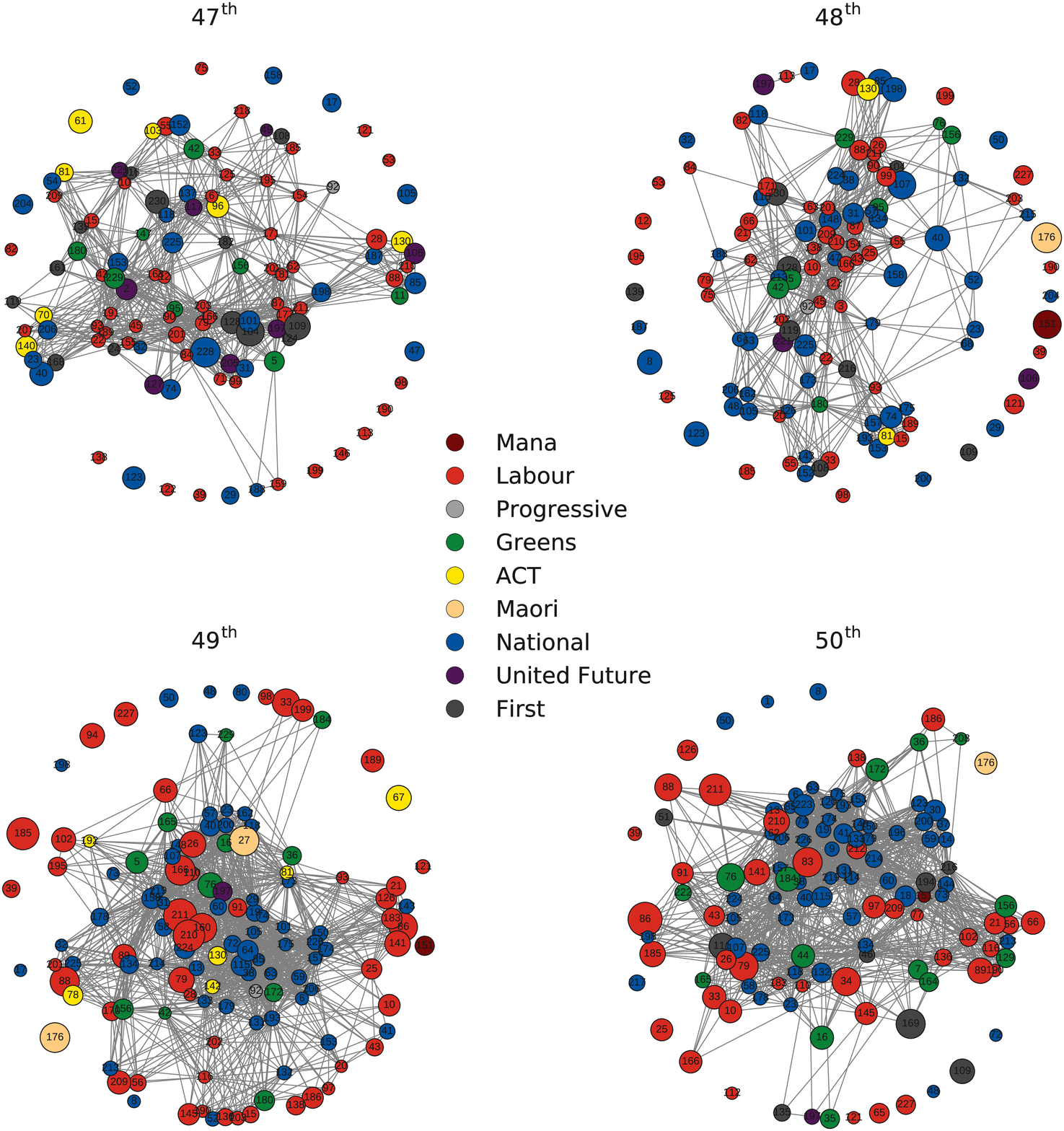}
\caption{(Color online) MP networks for parliaments $47^{\mathrm{th}}$ to $50^{\mathrm{th}}$ corresponding to the MP-projection of the original bipartite graph. Node size is proportional to the total number of words spoken by each MP over the course of the parliamentary term. Node labels identifying MPs are provided in \ref{longtable1}, Supporting Information.}
\label{networks}
\end{figure}

The MP-projected networks for the $47^{\mathrm{th}}$ to $50^{\mathrm{th}}$ parliaments resulting from the process described above are shown in Fig. \ref{networks}. The community structure \cite{fruchterman1991graph} in these networks is visible, as is the party make-up of these communities. Table \ref{numberMPs} shows the number of MPs per party present in each of these four networks.

\begin{table}[h!]
\vspace{0.2cm}
\centering
\caption[\label{numberMPs}]{Number of Members of Parliament in the three largest parties (as of the 50$^{\mathrm{th}}$ Parliament) for each complete parliamentary term studied. \label{numberMPs}}
\vspace{0.2cm}
\begin{tabular}{ccccc}%
\toprule
\bfseries Term &  \bfseries National  & \bfseries Labour & \bfseries Green & \bfseries Others\\
\midrule
\csvreader[
    late after line=\\,
    late after last line=\\ \hline,
    before reading={\catcode`\#=12},
    after reading={\catcode`\#=6}]%
    {MPs_perparty_perterm.csv}{1=\one,2=\two,3=\three,4=\four,5=\five}{\one & \two &  \three & \four & \five}
\end{tabular}
\end{table}

\begin{figure}[!h]
\centering
\includegraphics[width=0.75\textwidth]{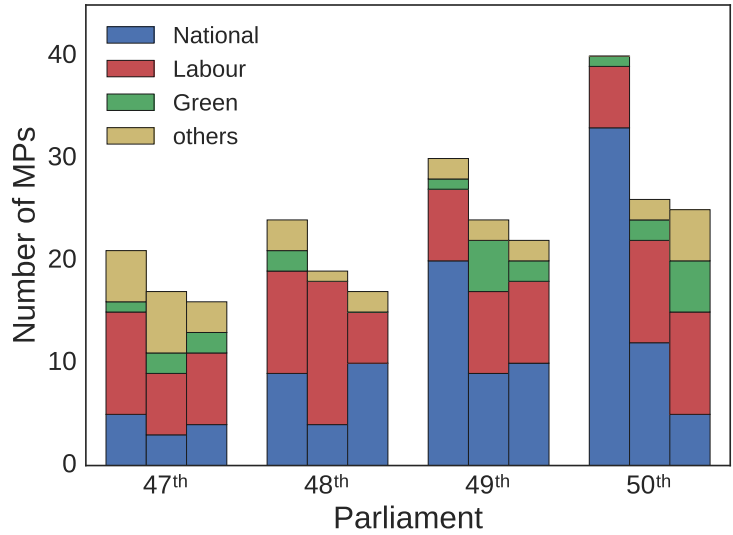}
\caption{Size and party composition of the three largest communities discovered in the MP-projected network for each of the parliaments examined.}
\label{commpart}
\end{figure}

\begin{figure}[!h]
\centering
\subfigure[\label{degtopic}]{\includegraphics[scale=0.3]{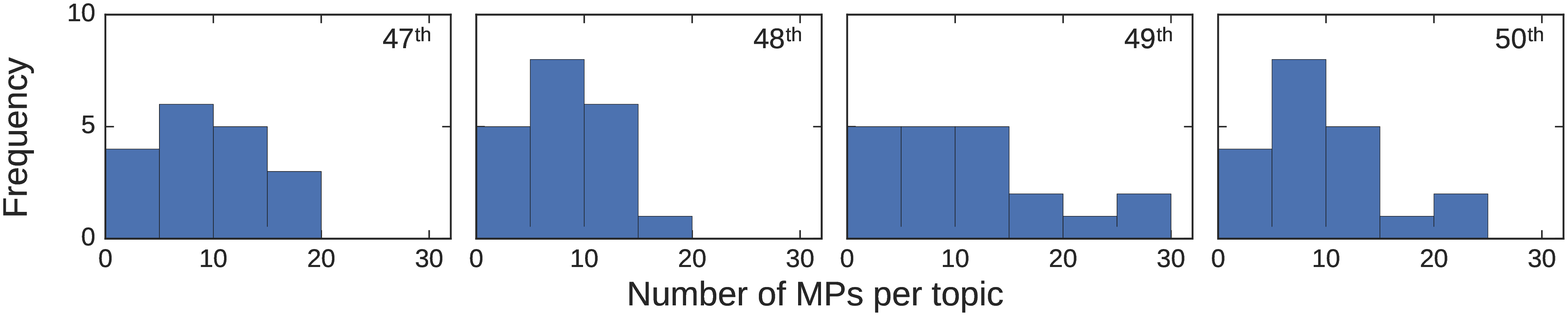} }
\subfigure[\label{degMPbi}]{\includegraphics[scale=0.3]{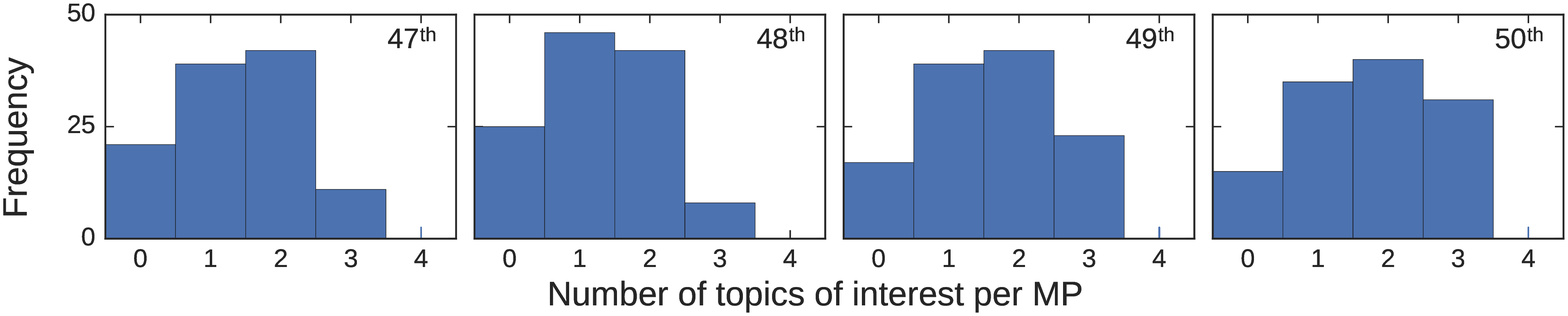} }
\subfigure[\label{degMPproj}]{\includegraphics[scale=0.3]{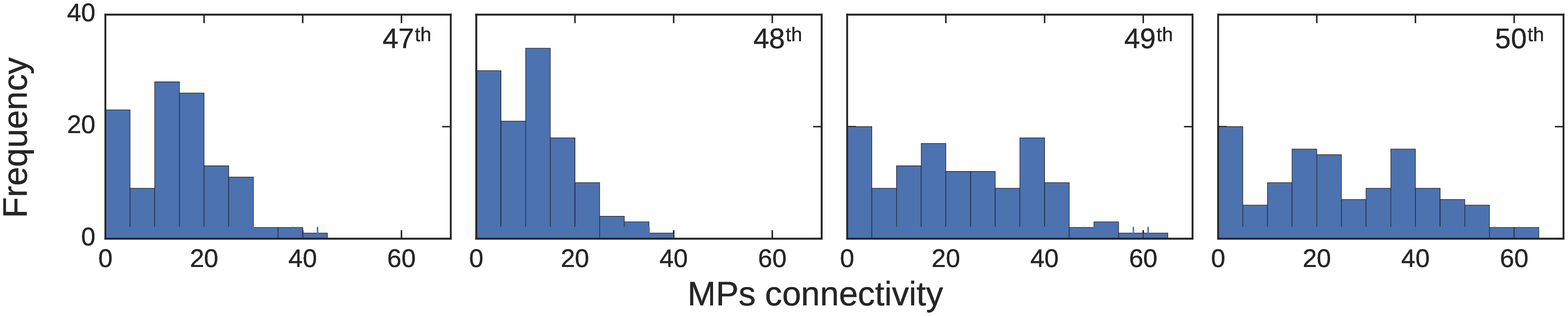} }
\caption{Frequency distributions of the (a) number of MPs linked to each topic in the original bipartite graph, (b) number of topics linked to each MP in the original bipartite graph, and (c) number of links each MP has to other MPs in the MP-projected network.}
\label{fig:degreedistr}
\end{figure}

\begin{figure}[!h]
\centering
\subfigure[\label{avdegree}]{\includegraphics[scale=0.3]{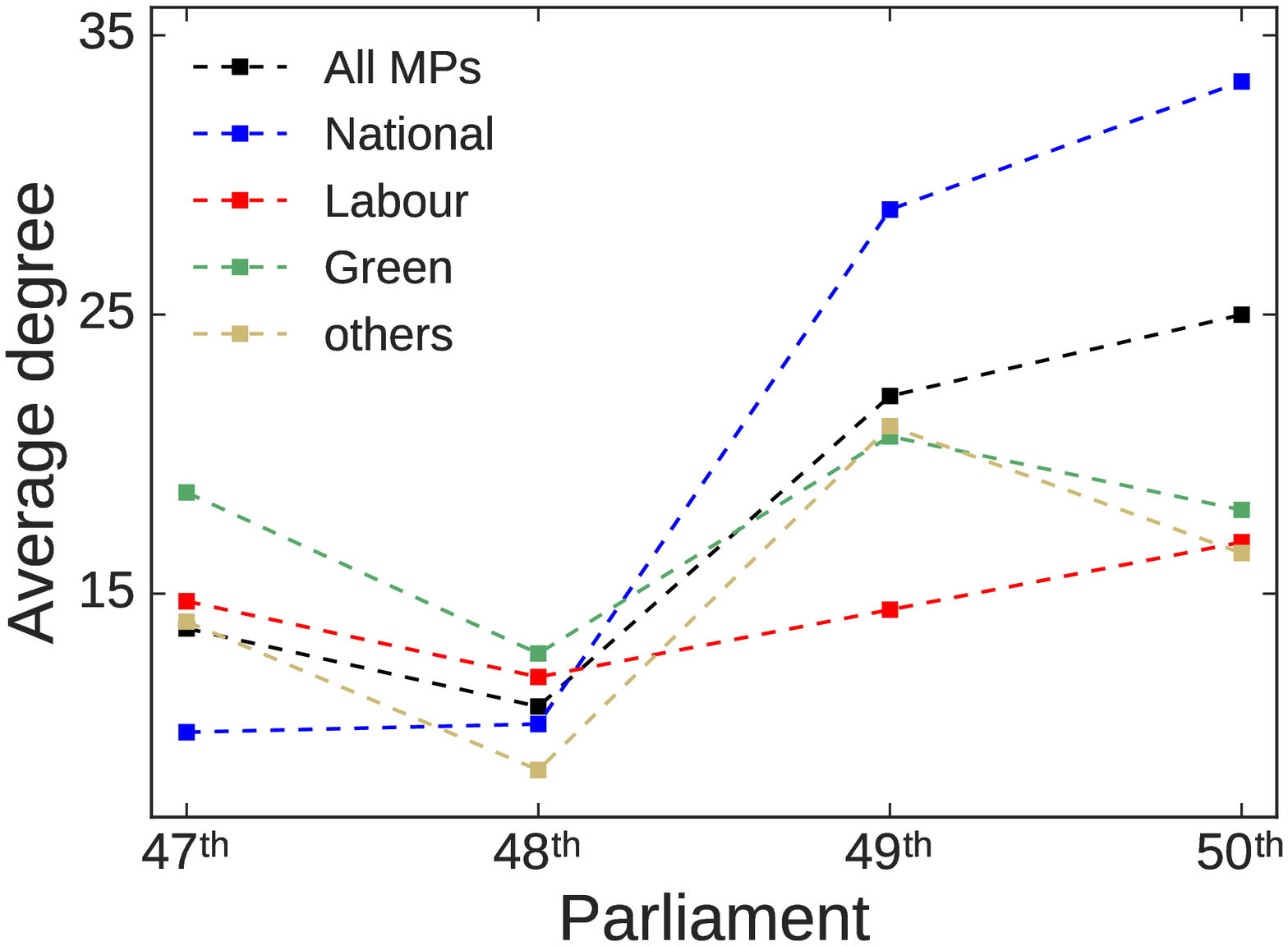} }
\subfigure[\label{homophily}]{\includegraphics[scale=0.3]{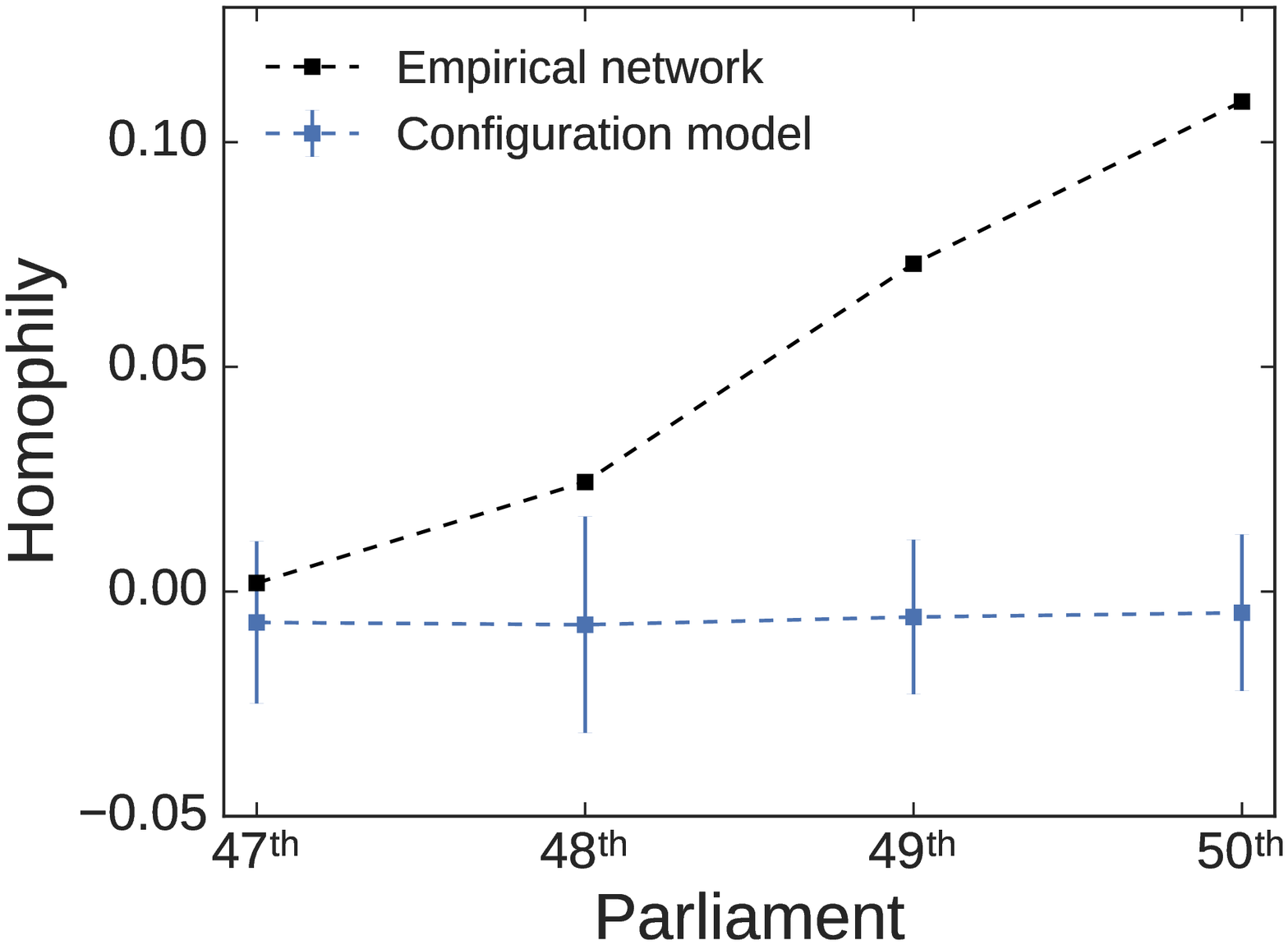} }
\caption{Left: Time series of the average degree of MPs in the MP-projected network, decomposed by party. Right: Time series of net homophily in the MP-projected networks for the empirical data and configuration model. It shows that if connections were random, the expected homophily would stay nearly constant and the network would be slightly dissortative. The results show average and standard deviation over 1000 runs.} 
\label{fig:features}
\end{figure}



Before examining the structure of the network, it is important to note that in all networks a number of MPs are not connected to any others. This is a reflection of our methodology. The 6.7\% threshold that is applied to filter topics may in fact remove all topics in the unlikely scenario that the MP in question talks about many topics in roughly equal proportion. MPs with diverse interests may not be identified as fitting into any particular community. 

A striking difference between the MP networks of the first two parliaments ($47^{\mathrm{th}}$ and $48^{\mathrm{th}}$) and last two ($49^{\mathrm{th}}$ and $50^{\mathrm{th}}$) is the party composition of visible communities. In the first two (Labour government, 2002-2008) the communities are quite party-wise diverse, while in the last two (National government, 2008-2014) there is a close-knit community made up of National Party MPs. This is corroborated by the three largest communities composition, for every term in our analysis, shown in Fig. \ref{commpart} \cite{blodel2008fast}. In the first two terms, we note heterogeneous, smaller core communities and the absence of a community that is much larger than the others. That changes for the last two terms, specially the last one, where we see one community much larger than the other emerging, dominated by MPs of the National Party. Another point worth noticing is the smaller presence of minor parties in the largest communities over time, ending with no presence whatsoever in the largest community of the $50^{\mathrm{th}}$ term network.

Also supporting this idea of a close-knit community made up of National Party MPs is Fig. \ref{homophily} that compares homophily between MPs (based on party affiliation) for the empirical data and configuration model networks. The latter is a random model in which we keep the same degree sequence of the empirical one and rewire the links. It shows that if connections were just random, then the expected homophily would be fairly constant and the network would be slightly dissortative. On the other hand, there is an increasing homophily of the empirical networks, meaning that MPs preferentially share interests with other members of their parties, particularly within the National Party. \\

The degree distributions in Fig. \ref{fig:degreedistr} also tell an interesting story. From Fig \ref{degtopic} we see that topics are attracting the interest of more MPs over time. Associate with that is Fig. \ref{degMPbi} that shows the distribution of topics that MPs spend the most time on during each government. Most MPs speak about 2-3 topics in large proportions over these periods, however for parliaments 47-49 we see a trend towards larger repertoires. 

From Fig. \ref{degMPproj}, we can see the Labour government in the $47^{\mathrm{th}}$ and $48^{\mathrm{th}}$ parliaments display a fairly sparse network, where most MPs share interests with less than twenty other MPs. The $49^{\mathrm{th}}$ and $50^{\mathrm{th}}$ parliaments appear to show the formation of a compact National Party group (visible in Fig. \ref{networks}) speaking about the same topics as many others within this group, such that members of this group in the $50^{\mathrm{th}}$ parliament are connected to at least 45 other MPs, most of whom are also members of this group.


\section{Discussion}
\label{sec:discussion}

Topic models provide a way to parse human speech and extract themes from large bodies of text that are often difficult and time consuming to analyze manually. In few cases is it more important to gather and process this information than in the speeches of those people that control the legislative and political direction of a country. Topic modeling is unlikely to replace traditional media analysis of political speech, however, here we have shown that it is a useful tool in examining larger themes and trends in political discourse. We were able to use topic modeling to track changes in the content of parliamentary speeches across time, and identify features in these time-series that correspond to particular issues or events.

In the time period examined, a number of large events influenced political discourse in New Zealand, such as the 2011 Christchurch earthquake, the global economic crisis and changes to local government with the creation of the Auckland `Super City' via the amalgamation of numerous smaller councils (see  Fig. \ref{fig:timeevolutionbyparty3}) as well as more recently, the housing crisis. Parliamentary discussions around all of these topics were identified, alongside more conventional themes such as the economy, the budget, and social welfare. 

Breaking down these topics by time and party shows the different emphases parties are putting on topics. For example, we can see that much of the discussion around the developing housing crisis has been pushed by Labour (see Fig. \ref{fig:timeevolutionbyparty1}) and, to a lesser extent, the Green Party, while the governing National Party showed little additional interest. Conversely, around the time of the economic downturn the National Party spent more of its time talking about economics. In other cases, such as the discourse surrounding the Christchurch earthquake and the governance of the Canterbury region (see Fig. \ref{fig:timeevolutionbyparty1}), increase in discussion was driven by all parties. Unsurprisingly, the party that spend the most of their time discussing the environment, was the Green Party. 

Some events are sufficiently large that it would be difficult for a political party to ignore them, such as the Christchurch earthquake which influences the Canterbury topic. Exogenous events such as these force politicians to comment, driving conversation across party lines. Topics where there are major differences in trends suggest endogenous drivers, where discussion is the result of conscious decisions by political parties. The National Party's apparent indifference to the increasingly vocal opposition parties' discussion around housing over the period examined would suggest that National were consciously choosing not to engage with this topic, while Labour and Green were consciously choosing to engage. 

A mixture of mechanisms could also drive changes in the level of discussion. The increase in the discussion of the economy by National appears to occur at the same time as the global financial crisis. It also coincides with the National Party taking power. The change in discussion could be attributed to the crisis, or it could be that the governing party simply talks proportionally more about the economy when they first enter office. Continuing this analysis past another change of government sometime in the future would allow us to identify the drivers of this type of pattern.  

From a basic analysis of the topic proportions we were able to identify an increase in the number of topics under discussion per MP, alongside an increase in the total number of words spoken per year. We can also observe that parties in opposition tend to talk more than parties in power. In the $47^{\mathrm{th}}$ and $48^{\mathrm{th}}$ parliaments, the National Party has a greater total word count. In the $49^{\mathrm{th}}$ and $50^{\mathrm{th}}$ after the National Party takes power, the Labour Party becomes the most vocal (see Fig. \ref{wordsparty})

Topic models also produce a natural bipartite network that can be decomposed into its projections and analyzed using standard network techniques. Without the use of sophisticated or computationally expensive methodologies we have shown that the networks resulting from a topic model can display useful information such as community structure and interpretable degree distributions.

Much of the popular political analysis since the current National government took power in 2008 (the $49^{\mathrm{th}}$ parliament) has noted the factionalism with the major opposition party, Labour. The community structure we have inferred (Fig. \ref{commpart}) supports this more traditional analysis, with Labour MPs speaking more, on disparate topics while National MPs largely kept to a smaller number of topics. Extracting and examining the top three communities for each parliament we can see the National Party coming to dominate discussion within the core communities in the $49^{\mathrm{th}}$ and $50^{\mathrm{th}}$ parliament. We also see a gradual decline in the participation of smaller parties in these largest communities identified, in particular, having no influence in the largest community in the $50^{\mathrm{th}}$ parliament. 

Fig. \ref{fig:degreedistr} shows the gradual increase in the number of topics discussed by MPs over time (see Fig. \ref{degMPbi}), indicating decreasing topic specialization by individual MPs in their parliamentary speeches. This has resulted in MPs becoming more highly connected over time (see Fig. \ref{degMPproj}).  At the same time, the average degree of National MPs has increased significantly (\ref{avdegree}). When considered in light of the communities shown in \ref{commpart} this analysis suggests a widening of political discourse in New Zealand with opposition parties talking about a greater number of topics, and the development of tight knit communities mostly consisting of government MPs talking about a smaller range of topics.




\section*{Acknowledgments}
The authors would like to thank Te P\={u}naha Matatini for funding this project. 
\section*{Author contributions}
The four authors designed the study, contributed to the analysis of the results and to the writing of the manuscript.

%
%
%

\pagebreak


\appendix

\counterwithin{figure}{section}
\counterwithin{table}{section}

\section{Supporting Information}

\subsection*{Codes and names of Members of the Parliament}

\begin{longtable}{lr|lr} 
\caption[]{Members of the Parliament and their respective codes in the networks. Codes that are missing in this table is due to the fact that some speakers don't show in the network. This is either because they were invited speakers (not a MP) or because the MP had spoken too few words (below ten thousand words) during the entire term.\label{mpcodestable}}\\
\toprule
\bfseries Code & \bfseries MP &  \bfseries Code  & \bfseries MP\\
\midrule \endfirsthead \endhead
\bottomrule \endfoot 
\csvreader[
    late after line=\\,
    late after last line=,
    before reading={\catcode`\#=12},
    after reading={\catcode`\#=6}]%
    {mpcodes.csv}{1=\one,2=\two,3=\three,4=\four}{\one & \two &  \three & \four}
\label{longtable1}
\end{longtable}
\pagebreak

\subsection*{Topics identified and keywords}

\begin{longtable}{p{.20\textwidth} | p{.735\textwidth} } 
\caption[]{Topics identified and their respective key words from Mallet \\ 
* Topics that were removed from the networks. \\
** Sometimes just part of Māori words appear due to accent marks. Those are not readable in the format used in Mallet (e.g. 'whānau' would appear as 'wh' only). Speeches in Māori are translated to English, i.e. the corpus in this work contains these speeches in their translated versions. Therefore, this topic needed to be removed from the network.
\label{keywordstable}}\\
\toprule
\bfseries Topic  & \bfseries Key words \\
\midrule \endfirsthead
\endhead
\bottomrule \endfoot
\csvreader[
    late after line=  \\\hline ,
    late after last line= ,
    before reading={\catcode`\#=12},
    after reading={\catcode`\#=6}]%
    {keywords.csv}{1=\one,4=\four}{\one & \four}
\end{longtable}
\pagebreak

\subsection*{Time series of topic proportions}

\begin{figure}[!h]
\vspace{0cm}
\centering
\includegraphics[width=1.0\textwidth]{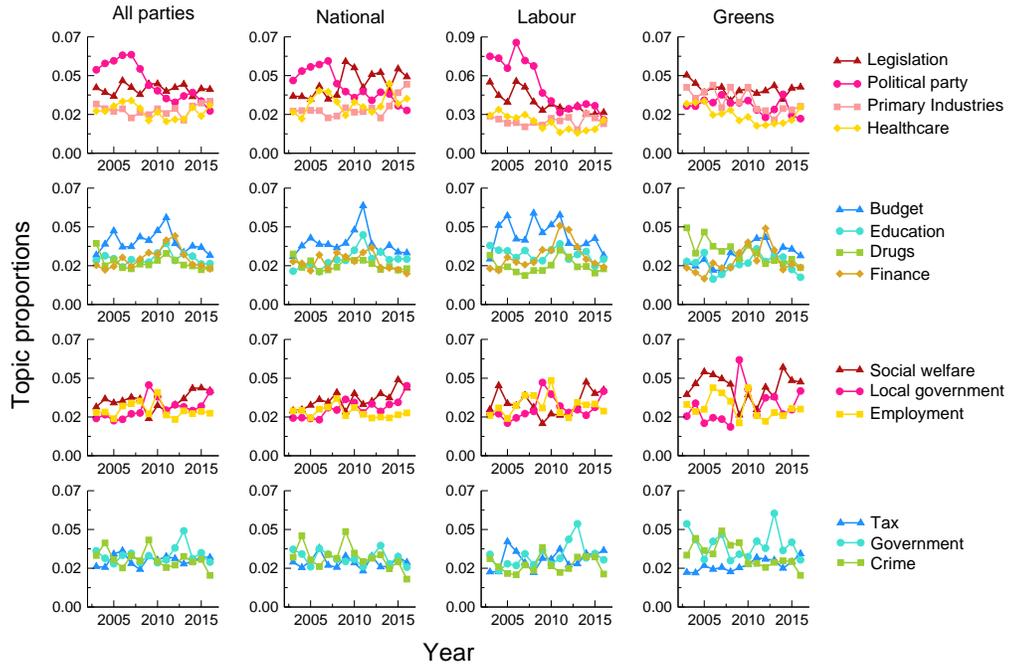}
\vspace{0mm}
\caption{Time evolution of topic proportions, as discussed at Parliament and its decomposition by party.} 
\label{fig:timeevolutionbyparty3}
\end{figure}




\subsection*{Histogram of number of words spoken by MP}

\begin{figure}[!h]
\vspace{0cm}
\centering
\includegraphics[width=0.7\textwidth]{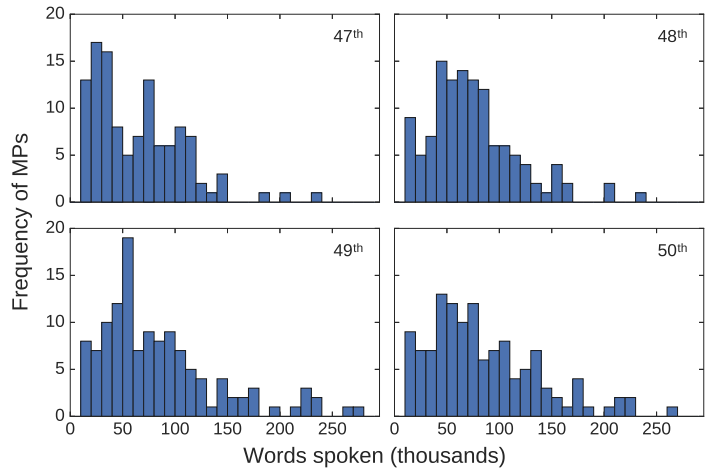}
\vspace{0mm}
\caption{Histogram of number of words spoken by MP during the four parliaments.} 
\label{fig:wordsbyMP}
\end{figure}

\end{document}